\documentclass[final, 11pt]{IEEEtran}
\IEEEoverridecommandlockouts
\usepackage{cite}
\usepackage{amsmath,amssymb,amsfonts}
\usepackage{algorithmic}
\usepackage{graphicx}
\usepackage{subcaption}
\usepackage{textcomp}
\usepackage{xcolor}

\def\BibTeX{{\rm B\kern-.05em{\sc i\kern-.025em b}\kern-.08em
    T\kern-.1667em\lower.7ex\hbox{E}\kern-.125emX}}
\begin{document}

\title{Enhancing Explainability in Solar Energetic Particle Event Prediction: A Global Feature Mapping Approach
}


\author{\IEEEauthorblockN{Anli Ji\textsuperscript{1}, Pranjal Patil\textsuperscript{2}, Chetraj Pandey\textsuperscript{3}, Manolis K. Georgoulis\textsuperscript{4}, Berkay Aydin\textsuperscript{2}}
\\
\IEEEauthorblockA{\textit{\textsuperscript{1}Dept. of Computer Science, California State University, Fullerton, CA, USA} \\
\textit{\textsuperscript{2}Dept. of Computer Science, Georgia State University, Atlanta, GA, USA} \\
\textit{\textsuperscript{3}Dept. of Computer Science, Texas Christian University, Fort Worth, TX, USA} \\
\textit{\textsuperscript{4}Applied Physics Laboratory, Johns Hopkins University, Laurel, MD, USA} \\
\textit{\textsuperscript{1}Corresponding Author: anli.ji@fullerton.edu} 
}
}

\maketitle

\begin{abstract}
Solar energetic particle (SEP) events, as one of the most prominent manifestations of solar activity, can generate severe hazardous radiation when accelerated by solar flares or shock waves formed aside from coronal mass ejections (CMEs). However, most existing data-driven methods used for SEP predictions are operated as black-box models, making it challenging for solar physicists to interpret the results and understand the underlying physical causes of such events rather than just obtain a prediction. To address this challenge, we propose a novel framework that integrates global explanations and ad-hoc feature mapping to enhance model transparency and provide deeper insights into the decision-making process. We validate our approach using a dataset of 341 SEP events, including 244 significant ($\geq$10 MeV) proton events exceeding the Space Weather Prediction Center S1 threshold, spanning solar cycles 22, 23, and 24. Furthermore, we present an explainability-focused case study of major SEP events, demonstrating how our method improves explainability and facilitates a more physics-informed understanding of SEP event prediction.
\end{abstract}


\section{Introduction}
\begin{figure*}[tb!]
  \centering
  \begin{subfigure}{0.25\textwidth}
    \includegraphics[width=\linewidth]{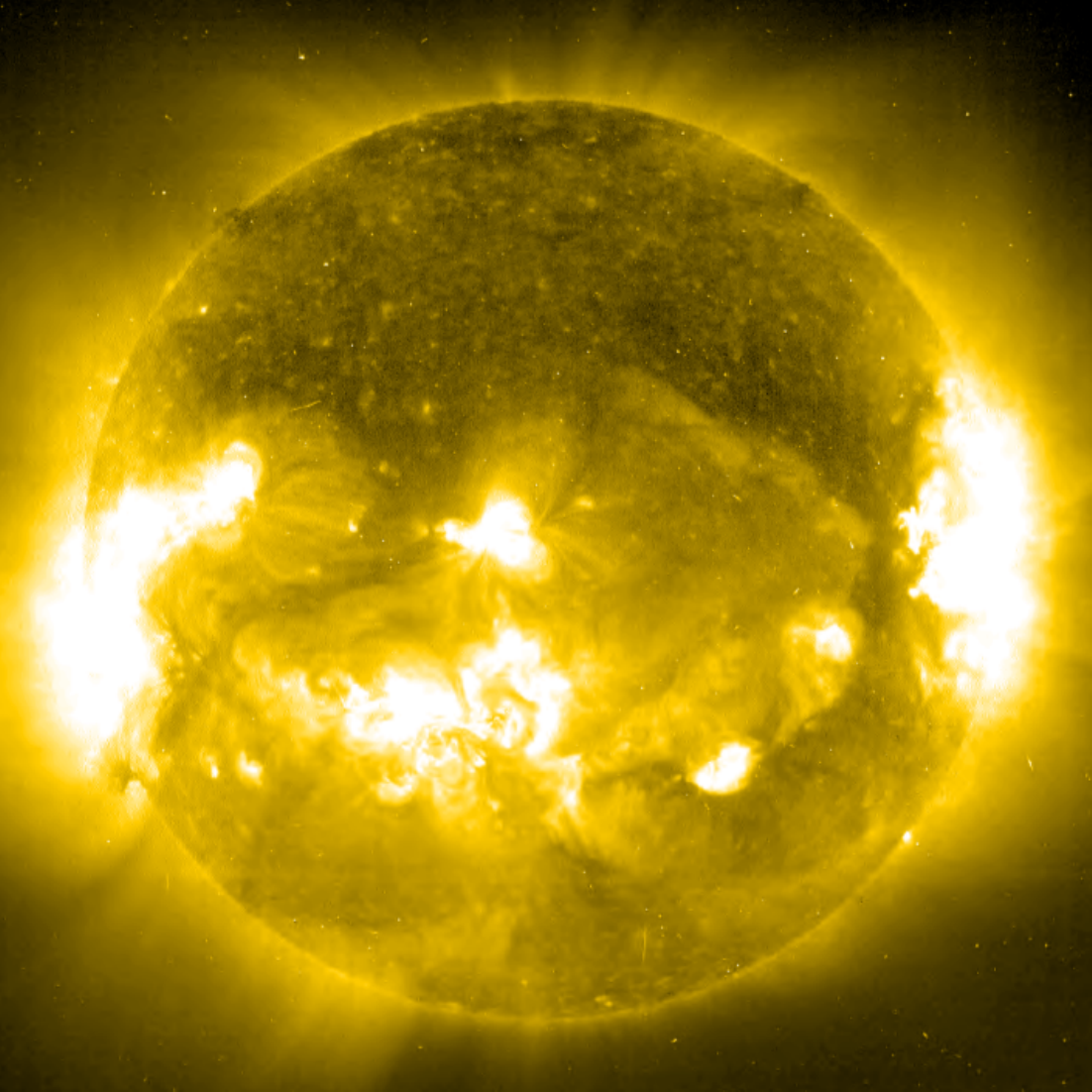}
    \caption{}\label{fig:1a}
  \end{subfigure}%
  \hspace*{\fill}   
  \begin{subfigure}{0.25\textwidth}
    \includegraphics[width=\linewidth]{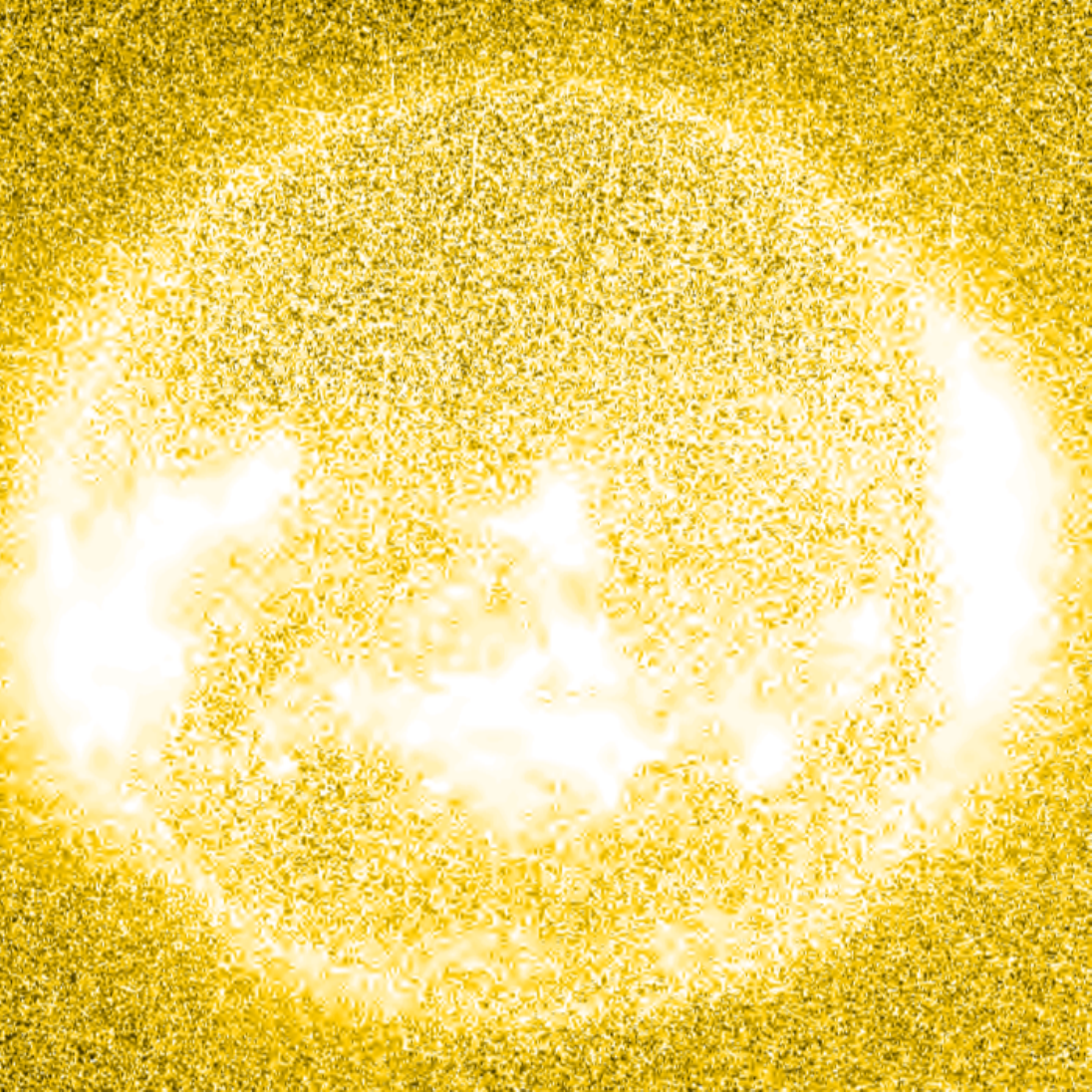}
    \caption{}\label{fig:1b}
  \end{subfigure}%
  \hspace*{\fill}   
  \begin{subfigure}{0.5\textwidth}
    \includegraphics[width=\linewidth]{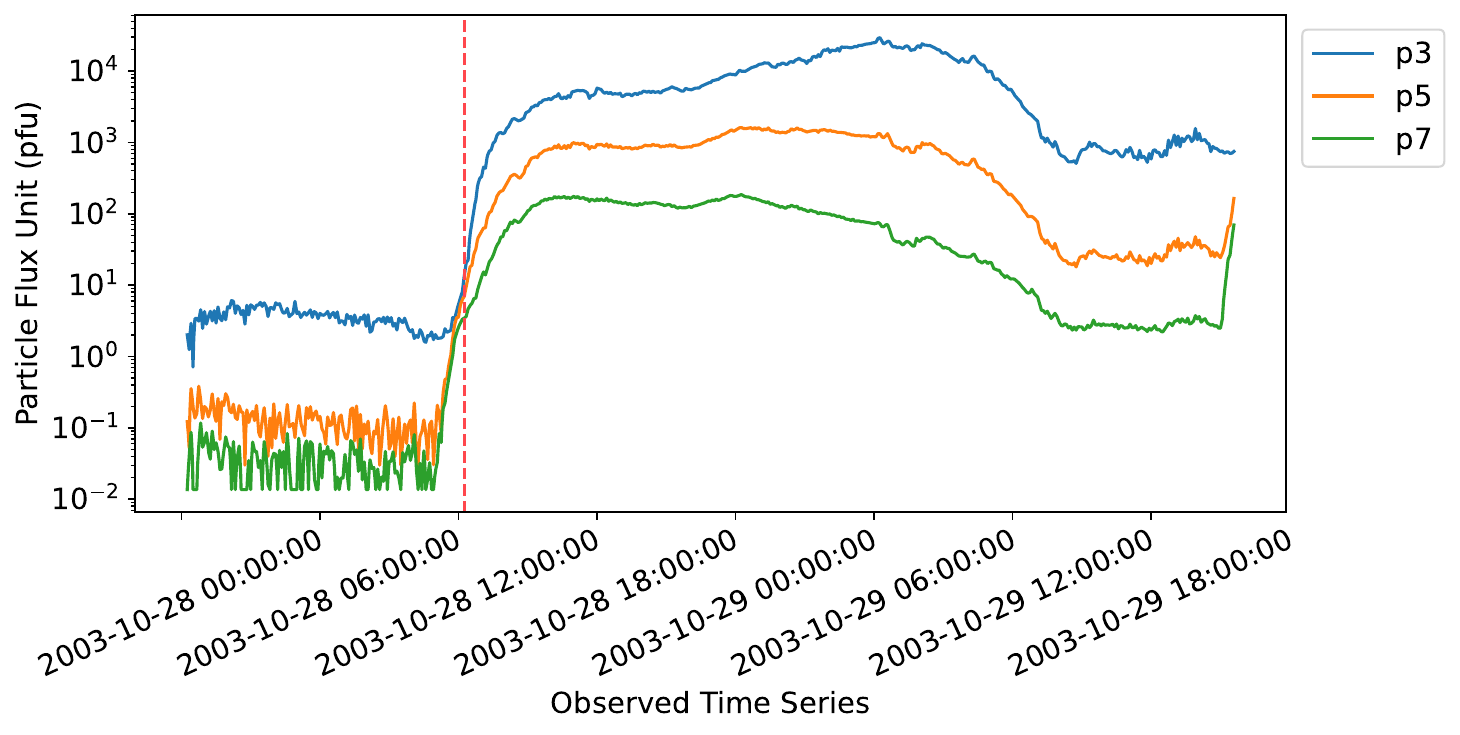}
    \caption{}\label{fig:1c}
  \end{subfigure}%
  \caption{Solar Energetic Particle (SEP) event during the Halloween storms, captured by Extreme Ultraviolet Imaging Telescope (EIT) onboard Solar and Heliospheric Observatory (SOHO) from (a) 2003-10-28T07:06 (onset) to (b) 2003-10-28T22:11 (peak). (c) Time series plot occurred on 2003-10-28T11:35 (UT) shown in log scale within GOES P3($\geq$10 MeV), P5($\geq$50 MeV), and P7($\geq$100 MeV) integral proton channels.} \label{fig1}
\end{figure*}

Solar Energetic Particle (SEP) events are characterized as one of the most severe space weather phenomena that release high-energy particles (e.g., protons, electrons, and heavy ions) from the Sun. Often triggered by solar flares or coronal mass ejections (CMEs), these phenomena are capable of emitting high levels of magnetic radiation, leading to potential risks in space missions and technological infrastructure. Severe impacts include damage to orbiting electronics, disruption of high-frequency radio communication systems and GPS navigation networks, and even the danger to spacecraft astronauts' health \cite{Iucci2005} \cite{Kim2011} \cite{Jones2005} \cite{Eastwood2017}. An illustrative example of a SEP event that occurred during the Halloween storms is shown in Fig.~\ref{fig1} with a corresponding time series of GOES integral proton channels. 

Despite its importance, the reliable forecasting of SEP events remains challenging. One primary reason is data scarcity, since SEP events are relatively rare (i.e., especially the strong events with significant potential impacts near Earth), limiting the availability of comprehensive observational data required for training models. In the most extreme cases, these high-energy particles can travel to Earth within tens of minutes, making real-time prediction and issuing timely warnings particularly challenging. In addition, the underlying physical fundamentals of generating such events are complex and have not yet been fully understood \cite{Vlahos2019} due to the combined physics of particle injection, acceleration, and transport being highly complex \cite{Anastasiadis2019}. Another factor comes from the fact that the observed characteristics of SEP events vary depending on their location within the heliosphere. Specifically, the intensity, temporal profile, particle composition, and spectral distribution of each SEP event varies when measured from different spacecraft or observatories positioned at different radial distances relative to the Sun \cite{Cliver2008}. This variability arises from differences in magnetic field connectivity between the observer and the source region of the SEP as well as differences in the geometry of the interplanetary magnetic field and solar wind conditions in which the particles are scattered as they propagate. As a result, accurate and generalized modeling of SEP events becomes challenging since predictions based on observations from one location may not necessarily be applicable or accurate for another vantage point elsewhere in the heliosphere.

Machine learning (ML) has recently succeeded in introducing new possibilities for SEP prediction, particularly in exploring hidden patterns (often non-linear relationships) in multidimensional observational data. Learned models are able to capture relationships across various variables (i.e., physical parameters), such as flare intensities, CME properties, and solar radio flux variations, therefore potentially outperforming many traditional empirical or physics-based models \cite{Whitman2023}. However, ML-based models come with their own challenges, mainly related to their inherent "black-box" nature. When generating predictions, they often lack transparency on how specific inputs derived into the outcomes. This lack of interpretability significantly limits the operational usability of SEP forecasts, especially given the high-stakes implications for satellite operations.

To tackle this issue, we present a novel framework that integrates feature mapping and global explanations to enhance the interpretability of ML-based SEP event predictions. Unlike local explanation methods that interpret predictions for individual instances (e.g., single predictions), global feature mapping provides a holistic overview, highlighting patterns and relationships that the model has learned across all available data. 
Our research intends to create a more interpretable and reliable SEP forecasting model, improving both its scientific implications and practical applicability within the broader context of solar physics. 

The rest of the paper is organized as follows: Section \ref{related_work} provides an overview of existing research and methodologies for SEP event forecasting. In Section \ref{methodology}, we introduce our proposed method, detailing the sliding window technique for extracting features from SEP time-series data. Section \ref{experiments} describes the experimental design, including data preparation, model implementation, and evaluation procedures. Lastly, Section \ref{conclusions} summarizes the key findings of this study and outlines potential directions for future research.

\section{Related Work}\label{related_work}

Traditional methods for SEP forecasting, as classified in \cite{Whitman2023}, can be divided into three main categories: empirical models, physics-based models, and hybrid methods. On the downside, empirical models are based on historical data and statistical relationships, making them sensitive to inaccuracies when encountering new or unprecedented information. Physics-based models tend to mimic the underlying physical processes involved, which are often computationally expensive and require a large amount of input parameters. Hybrid models combine the above-mentioned aspects and focus on balancing the trade-offs between accuracy, computational efficiency, and general applicability. The strength of such models compared to empirical methods comes from their capability of handling historical information, allowing them to learn complex, nonlinear relationships between input variables and SEP occurrence without the need for a priori incorporation of any physical knowledge.

More recently, researchers have begun exploring machine learning-based techniques as a potential avenue toward SEP forecasting. Typically, the built ML models involve utilizing training subsets (for identifying predictive relationships) and validation subsets (to assess forecasting performance). The prediction outcomes can be probabilistic, categorical, binary, or deterministic \cite{Whitman2023}. Examples include the binary classification of SEP events from GOES X-ray and proton channel observations \cite{Rotti2024} \cite{Boubrahimi2017} \cite{9750381}, solar radio flux data analysis using neural networks and genetic algorithms \cite{Kim2018}, and predictions based on the properties of solar flares and CMEs using ensemble methods \cite{Huang2012} \cite{9377906} \cite{aji2023}. While solar flares themselves do not directly cause SEPs, they are often associated with the acceleration of energetic particles. In fact, solar flares can trigger the acceleration of electrons and protons that are released during the flare, contributing to SEP events \cite{10302639} \cite{Pandey2023}. For solar flare prediction, both full-disk \cite{9671322} \cite{Pandey2022} \cite{10722839} and active region-based \cite{9378006} \cite{10460016, Hong2023, 10431579} \cite{9750381} approaches have shown significant impacts by utilizing derived time series features.

Regardless of their predictive capabilities, ML methodologies are often deployed as opaque systems (i.e., "black-box"), which poses challenges when physical interpretability is crucial \cite{pandey2023interpretable}. Complex models like deep neural networks, while capable of achieving high performance, usually have a hard time explicitly tracking the relationships between inputs and predicted outcomes due to their multi-layered, nonlinear structure. Moreover, SEP prediction normally involves high-dimensional data from multiple instruments and locations. The intricate interactions between these multi-dimensional data points are often abstracted within model architectures, making it challenging to explain the predictions in physically meaningful terms. In high-stakes domains, where decisions carry significant consequences, model interpretability is not merely a desirable characteristic but a critical prerequisite. The capacity to explain the rationale behind a model's output is indispensable for fostering a desire system and for the validation of the system's underlying logic.

Despite ongoing efforts, explainability remains a challenge in SEP event prediction. Existing methodologies mostly bring partial insights among parameters, and not all machine learning techniques integrate easily with current frameworks. Moreover, the rarity and intrinsic complex nature of SEP events further complicates the generation of reliable and physically meaningful explanations. Typically, explainability in SEP forecasting involves understanding the relationship between observational input variables (i.e., solar flare intensity, CME speed, magnetic connectivity, and solar radio flux measurements) and the predicted occurrence, intensity, or other characteristics of SEP events.

\section{Methodology}\label{methodology}

\subsection{\textbf{Sli}ding Window \textbf{M}ultivariate \textbf{T}ime \textbf{S}eries \textbf{F}orest}
The sliding-window multivariate time series forest (Slim-TSF) is an early fusion, interval-based ensemble classification method proposed in \cite{Ji2024}. It employs a multi-scale sliding windows approach that makes use of interval-based features extracted from all univariate time series. Such interval features calculate the statistical characteristics of intervals, including mean, standard deviation, and slope (as outlined in Eq.\ref{eq1:mean}, Eq.\ref{eq2:std}, and Eq.~\ref{eq3:slope}) as well as additional transformed features (i.e., maximum, minimum, and mean) comprising an additional localized pooling procedure used on the individual interval features extracted from a set of consecutive intervals obtained after the sliding window operation. To handle the multivariate nature of the time series, a Random Forest classifier is trained on the complete set of features extracted from all variables. The two groups of features serve as the foundation for creating this Slim-TSF classifier, which captures both the individual behaviors and interactions among variables within each interval.

\begin{equation}\label{eq1:mean}
f_{mean}(T^p_{s, e}) = \frac{\sum_{i=s}^{e} v_i}{s - e+1}
\end{equation}

\begin{equation}\label{eq2:std}
    \begin{split}
    f_{std}(T^p_{s, e}) = \sqrt{\frac{\sum_{i=s}^{e} (v_i-f_{mean}(T^p_{s, e}))^2}{s-l}} \text{,~~~~~~~~~~~}
    \end{split}
\end{equation}

\begin{equation}\label{eq3:slope}
f_{slope}(T^p_{s, e}) = \widehat{\beta} 
\end{equation}
where $\widehat{\beta}$ is the slope of the least squares regression line. An interval \(T_{s,e}\) is defined as a subsequence of the original time series $T$ such that \(<(t_s, v_s), (t_{s+l}, v_{s+l}),..., (t_e, v_e)>\) with starting point $s$ and ending point $e$ where \(1 \leq s < e \leq n\)

\subsection{Global Explainability}
In our approach to applying Slim-TSF to SEP event prediction, we extend the methodology to improve model transparency with global feature explainability. From the global perspective, we identify and rank the features based on importance, which is determined by how effectively each feature reduces the overall predictive uncertainty or impurity across the trees (measured by the Gini impurity metrics, shown in Eq.~\ref{eq:gini}). 

\begin{equation}\label{eq:gini}
Gini(D) = 1-\sum_{i=1}^{k}p_i^2
\end{equation}
where the probability of samples belonging to class $i$ at a given node can be denoted as $p_i$. Specifically, the importance of a feature is calculated by averaging its impurity reduction across all trees in the ensemble. A higher average reduction indicates greater significance. 

Additionally, instead of limiting our selection to only the highly ranked features from individual experiments, we implement a global mapping approach that aggregates the importance of all individual features. This approach relies on the cumulative outcomes of the entire bootstrapping process. Specifically, we repeatedly subsampled our dataset with replacement to create multiple instances. At each bootstrap iteration, the model produces a ranking of features according to their contribution to predictive performance. We record both the magnitude of each feature’s importance and its frequency of selection throughout the subsampling process. The cumulative aspect is achieved by systematically collecting the feature-importance information generated at each iteration of the subsampling process and then aggregating these results into a unified representation. 

By summing the importance scores across all iterations, we construct a cumulative importance distribution that reflects global relevance rather than local variability. While these rankings can vary slightly from one subsample to another due to data perturbations, the key idea is that the truly influential features will appear consistently across a wide range of iterations. Features that contribute meaningfully in only a few subsamples will have low cumulative importance, whereas those that repeatedly emerge as significant across diverse subsets will be amplified in the aggregated outcome.


\section{Results and Analysis}\label{experiments}
\subsection{Data Collection}
SEP events are often associated with solar flares and CMEs. Their initial eruptions are triggered by sudden disruptions of the magnetic field close to and within the active regions of the Sun's atmosphere. This relationship makes it well expected to build predictive capabilities using parameters related to these precursor solar events. Our primary data source is chosen based on this characteristic from the recently published open-source GSEP dataset \cite{Rotti2022} (available on Harvard Dataverse \cite{GSEP2022}). It is built using interpolated and time-integrated particle and X-ray flux data from several Geostationary Operational Environmental Satellite (GOES) missions with one-minute averaged solar X-ray fluxes (1–8 Å wavelength) recorded by the onboard X-ray Sensor (XRS), specifically three energy channels: (1) protons having energies $\geq 10$ MeV (P3 channel), (2) protons with energies $\geq50$ MeV (P5 channel), and (3) $E\geq100$ MeV proton fluxes (P7 channel).

The dataset categorizes SEP events based on the source SEP intensity as "Strong," "Weak," and "NoEvent." According to their type, they are labeled "1," "0," and "-1," respectively. A strong SEP event corresponds when one observes proton fluxes in excess of 10 pfu within the GOES P3 channel, whereas a weak SEP event is characterized by proton fluxes that remain below 10 pfu but above background between $\geq0.1$ and $\le10$ pfu. In total, this dataset comprises 244 strong SEP events that clearly exceed the threshold of 10 pfu in the GOES P3 channel and 189 weak events observed in near-Earth space from 1986 to 2018. Additionally, the dataset includes time-series slices of GOES proton and X-ray fluxes for all the events, where each slice consists of a 12-hour observation window prior to the event onset time, and the peak flux period of events. A detailed description of dataset generation and available parameters can be found in \cite{Rotti2023}.

\subsection{Experimental Settings}
In supervised classification tasks, datasets with labeled samples are commonly divided into distinct subsets with knowledge of the included labels \cite{Hastie2009}. The extracted features are used to configure the parameters of the chosen algorithm in the training set, and the classifier's predictive performance on new data is determined using the testing set. Given our prediction task as a classification problem, we partition our dataset into two non-overlapping subsets: a training set (i.e., 996 samples) and a testing set (i.e., 922 samples). Similar but extending to the forecasting approach in \cite{Rotti2024_1}, we explore the model capabilities for different short-term prediction windows of 6, 8, and 10 hours, as well as lag windows of 5, 15, 30, 45, 60, 120, and 180 minutes. Formally speaking:

\textit{\textbf{Definition:}} Let $I_{o} = (t_{o_{start}}, t_{o_{end}})$ be the observation window, $I_{lag} = (t_{o_{end}}, t_{sep})$ be the lag window, where $t_{o_{start}}$ and $t_{o_{end}}$ are observation window start and end times, while $t_{sep}$ is the start time of a SEP event. The length of these windows are defined as $\delta_{I_o} = t_{o_{start}} - t_{o_{end}}$ and $\delta_{I_{lag}} = t_{sep} - t_{o_{end}}$. Considering an anchored SEP event start time, we use $\delta_{I_o} \in \{ 6_{hrs}, 8_{hrs}, 10_{hrs} \} $ and $\delta_{I_{lag}} \in \{5_{mins}, 15_{mins}, 30_{mins}, 45_{mins}, 60_{mins}, 120_{mins}, 180_{mins}\}$

In this case, short-term predictions allow for more immediate and actionable decisions, particularly in terms of necessary precautions, such as adjusting satellite operations or mitigating potential damage to power grids.

\subsection{Evaluation Metrics}
To evaluate the performance of our model, we employ a binary $2\times2$ contingency matrix along with various evaluation metrics and crucial forecast skill scores. Within these evaluation metrics, the positive class corresponds to the occurrence of strong SEP events (i.e., $\ge$10 pfu according to the NOAA Space Weather Prediction Center), while the negative class refers to relatively weaker SEP events (i.e., $\geq0.1$ and $\le10$ pfu). Therefore, true positives ($TPs$) denote cases where the model accurately predicts an event (positive class), and true negatives ($TNs$) refer to instances where the model correctly predicts a smaller or no event (negative class). In contrast, false positives ($FPs$) represent false alarms, occurring when the model predicts a non-event as an ongoing one, and false negatives ($FNs$) indicate misses, which happen when the model fails to predict an actual event. In terms of SEP forecasting, we employ several well-known skill scores.

TSS can be calculated as the disparity between the probability of detection (recall of the positive class) and the probability of false detection (POFD). This measurement is determined using the equation specified in Eq.~\ref{eq:tss}.

\begin{equation}
    \label{eq:tss}
        TSS = \frac{TP}{TP + FN} - \frac{FP}{FP + TN}
\end{equation}

In essence, TSS can be reformulated as the sum of the true positive rate (i.e., $TPR$) and the true negative rate (i.e., $TNR$), offset by 1 (i.e., $TPR+TNR-1$).
The general purpose of TSS is a good all-around forecast evaluation method, especially for evaluating scores among datasets with different imbalance ratios. 



The Heidke Skill Score is another score that assesses how much the forecast enhances compared to a random forecast. This score varies from -1 to 1, where a score of 1 signifies flawless performance (whereas -1 indicates an all-opposite incorrect performance) and 0 denotes a lack of skill. This score of 0 means that the forecast's accuracy is no better than a random binary forecast based on the provided class distributions. This metric can be computed using Eq.~\ref{eq:hss2}, where $P$ represents the sum of true positives ($TP$) and false negatives ($FN$), and $N$ corresponds to the sum of false positives ($FP$) and true negatives ($TN$) in the observed data.

\begin{equation}
    \label{eq:hss2}
        HSS = \frac{2 \cdot ((TP \cdot TN) - (FN \cdot FP))}{P \cdot (FN + TN) + N \cdot (TP + FP)}
\end{equation}

For an extreme prediction task such as SEP events, prioritizing one evaluation metric over another can be problematic due to the typically high class imbalance present in the related datasets (such as SFs, discussed in \cite{Azim2019, Ahmadzadeh2021}). In such cases, a single metric like the Composite Skill Score (CSS, provided in Eq.~\ref{eq:css}) that balances between can be beneficial.

\begin{equation}\label{eq:css}
CSS= 
\begin{cases}
    \sqrt{\theta},& \text{if } \theta\geq 0\\
    0,              & \text{otherwise}
\end{cases}
\end{equation}
where $\theta=TSS\times HSS$. The CSS calculates the geometric mean average of both TSS and HSS, which takes into account the effects of compounding forecast ability (i.e., TSS) and enhancement over random assessment (i.e., HSS). 

The Gilbert Skill Score considers the number of hits due to chance, which is given as the frequency of an event multiplied by the total number of forecast events. This score formula is given by Eq.~\ref{eq:gss}

\begin{equation}
    \label{eq:gss}
    \begin{split}
        GSS = \frac{TP-CH}{TP+FP+FN-CH} \text{,~~~~~~~~~~~~~~~} \\ 
        \text{~~~~~where } CH = \frac{(TP+FP) \times (TP+FN)}{TP+FP+FN+TN}
    \end{split}
\end{equation}

Another evaluation metric we include is the F1 score, which is the harmonic mean of precision and recall. Their formulas are provided in Eq.s~\ref{eq:precision_xm}, \ref{eq:recall_xm}, and \ref{eq:f1}.

\begin{equation}
    \label{eq:precision_xm}
        Precision = \frac{TP}{TP+FP}
\end{equation}

\begin{equation}
    \label{eq:recall_xm}
        Recall = \frac{TP}{TP+FN}
\end{equation}

\begin{equation}\label{eq:f1}
F_1 = \frac{2 \times Precision \times Recall}{Precision + Recall}
\end{equation}

Giving equal weight to both precision and recall, the F1 score is particularly useful when dealing with imbalanced datasets, where one class has significantly more instances than the other. 




 


\begin{figure*}[t!]
  \centering
  \begin{subfigure}{0.8\textwidth}
    \includegraphics[width=\linewidth]{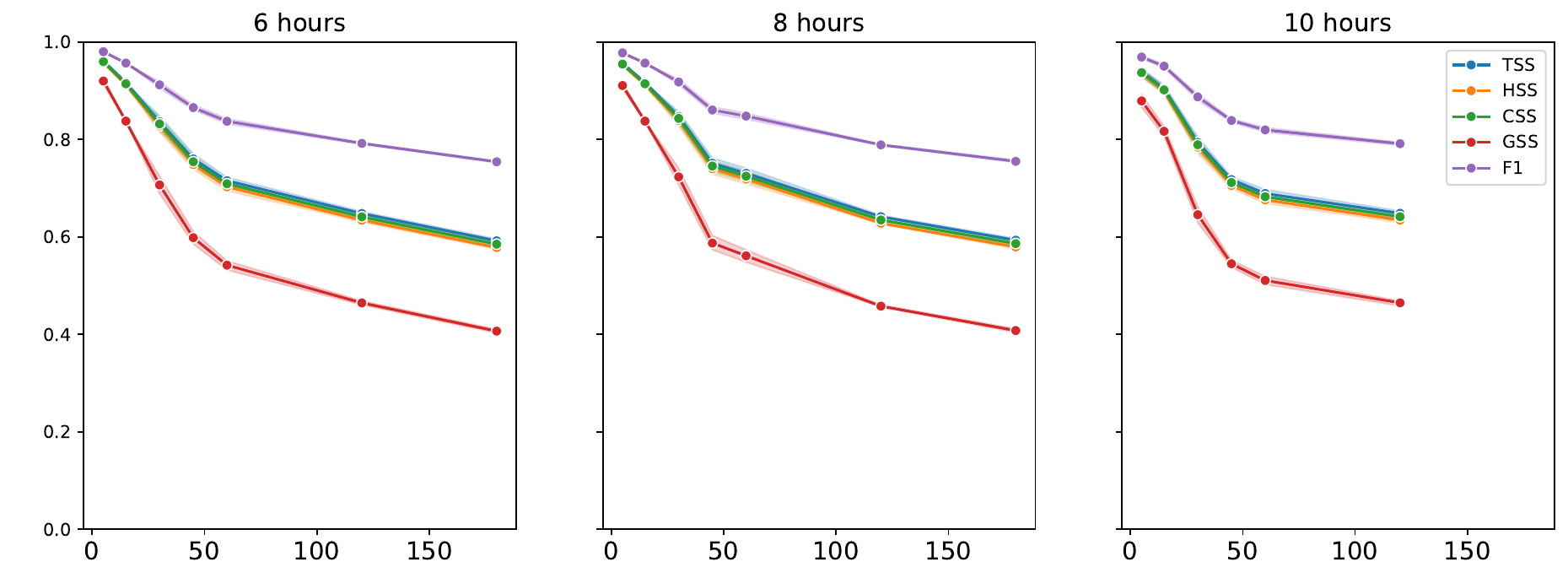}
    \caption{Strong vs. Weak}\label{fig:3a}
  \end{subfigure}%
  \\
  \begin{subfigure}{0.8\textwidth}
    \includegraphics[width=\linewidth]{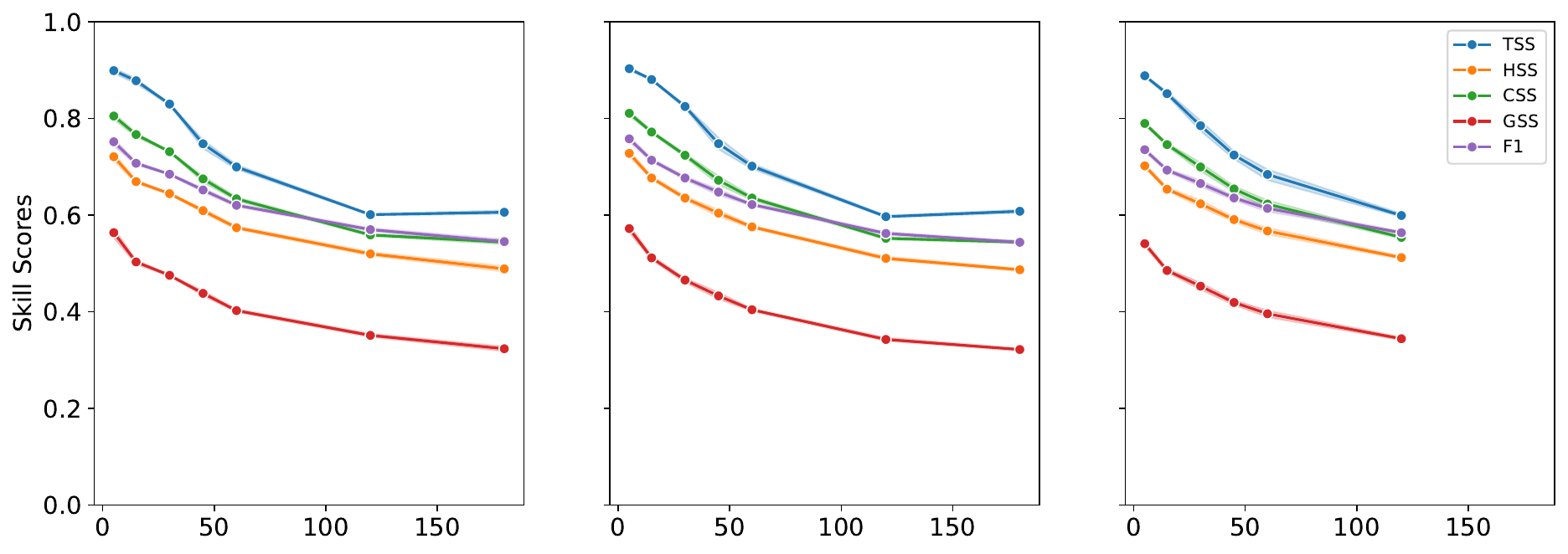}
    \caption{Strong vs. (Weak + NoEvent)}\label{fig:3b}
  \end{subfigure}%
  \\
  \begin{subfigure}{0.8\textwidth}
    \includegraphics[width=\linewidth]{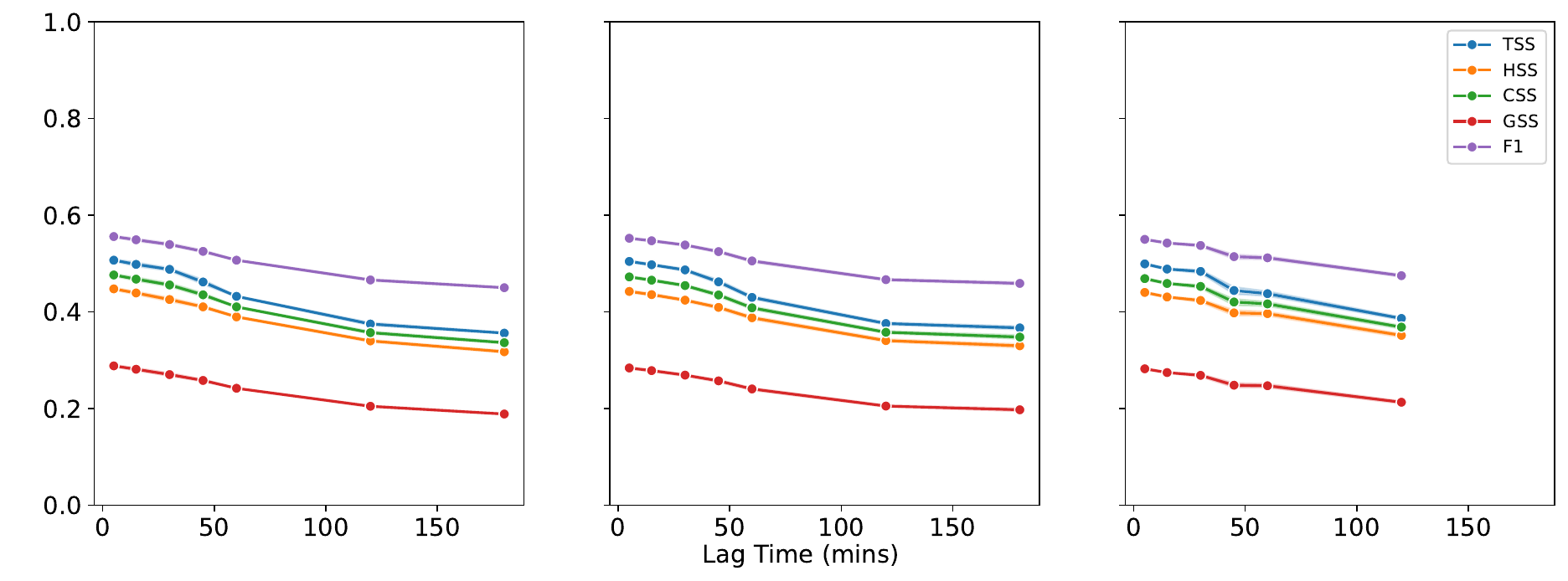}
    \caption{Event vs. NoEvent}\label{fig:3c}
  \end{subfigure}%
  \caption{Performance evaluation of the SEP event prediction model across three different classification scenarios: (a) Strong vs. Weak events; (b) Strong vs. Weak and No-event; and (c) Event vs. No-event at observation windows of 6, 8, and 10 hours and lag windows of 5, 15, 30, 45, 60, 120, and 180 minutes prior to the event onset time. Results were obtained by averaging across 10 bootstrapping runs.} \label{fig3}
\end{figure*}

\subsection{Model Performance}

To comprehensively evaluate our model's predictive capabilities, we conducted three distinct experimental scenarios reflecting different classifications relevant to SEP event prediction: (a) Strong vs. Weak Events, (b) Strong vs. Weak and No-event, and (c) Event vs. No-event. These scenarios capture increasingly broad definitions of SEP occurrence, from distinguishing event intensity levels to differentiating event occurrences from non-occurrences. As illustrated in Fig.~\ref{fig3}(a–c), the TSS, HSS, and GSS consistently indicate that the model's predictive accuracy and reliability reduced as the lag window increased. Specifically, for the Strong vs. Weak Events classifications (Fig.~\ref{fig:3a}), we observed that shorter lead times (lag closer to event onset) bring in significantly higher performance scores, highlighting the model's effectiveness from a short-term forecasting perspective. This suggests that distinct differences in SEP intensities become clearer closer to the event occurrence, enabling more accurate classification.

Similarly, in the Strong vs. Weak and No-event scenario (Fig.~\ref{fig:3b}), the skill scores remain relatively robust at shorter lag times but decrease as the time interval expands. This reflects that the model effectively distinguishes strong events from weaker or non-events when predictions are made closer to event initiation, highlighting the critical influence of immediate precursor parameters. Finally, the classification of Event vs. No-event (Fig.~\ref{fig:3c}) demonstrates generally lower scores overall, emphasizing the greater challenge in identifying subtle distinctions when weaker SEP occurrences are grouped with stronger events. The overall trend across these experiments strongly supports the conclusion that SEP event predictability is temporally dependent, emphasizing the importance of immediate observational data for accurate forecasting. These insights highlight the need for optimized, temporally precise feature extraction and interpretability techniques to improve prediction accuracy at critical lead times.

\subsection{Explainability Enhancement}

\begin{figure*}[!ht]
\includegraphics[width=0.7\textwidth]{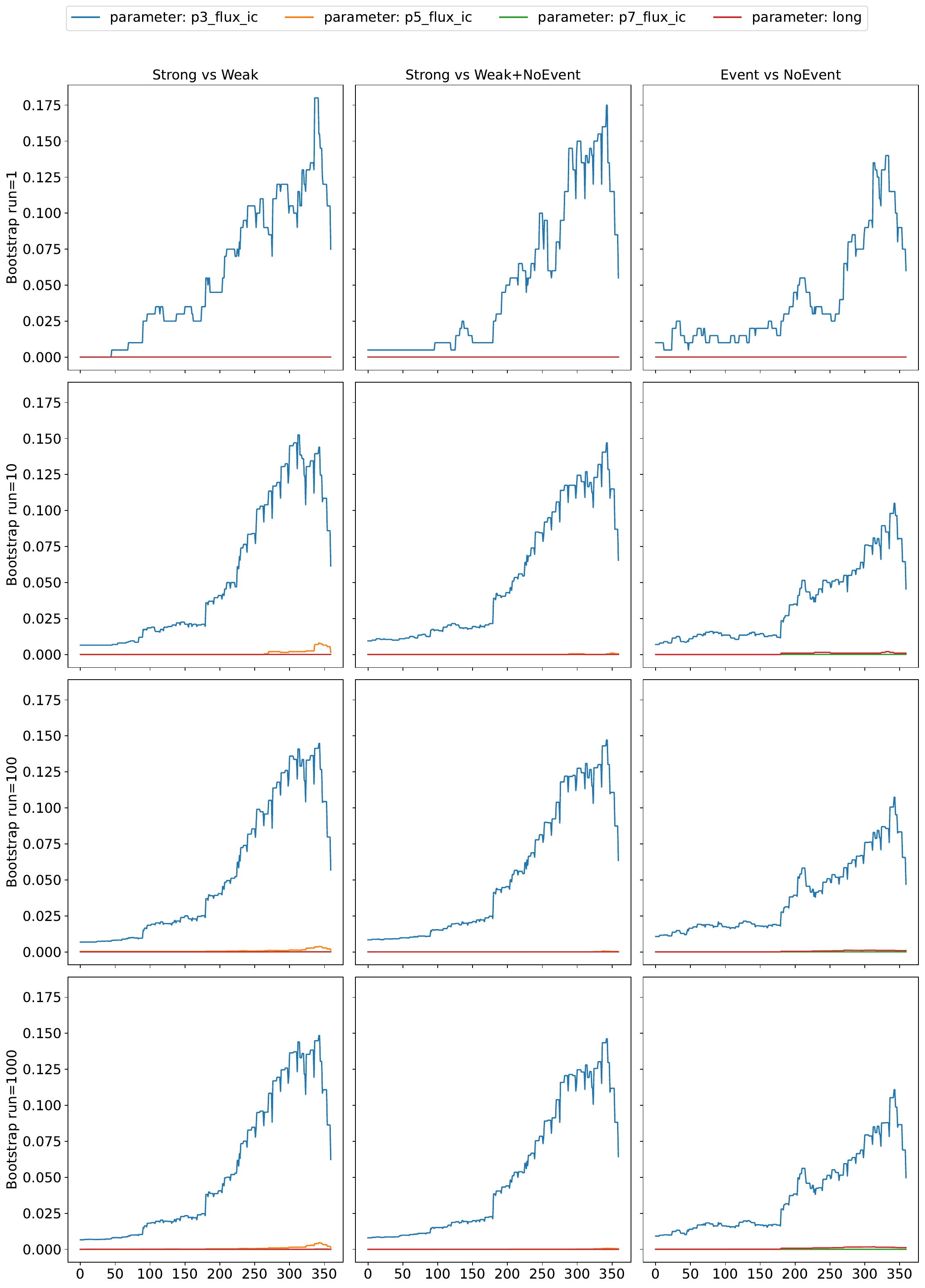}
\centering
\caption{Global explainability analysis of feature importance across flux channels (P3, P5, and P7), shown for bootstrap iterations of 1, 10, 100, and 1000 (top to bottom). Each subplot represents the integrated proton flux data across varying energy thresholds, providing insights into the temporal evolution of SEP events.} \label{fig4}
\end{figure*}

Fig.~\ref{fig4} illustrates the results of a global explainability analysis, highlighting the importance and contributions of various proton flux parameters (P3, P5, and P7 channels) used for predicting SEP events. Each subplot shows the relative importance of these parameters at different scales or conditions, providing insights into their predictive influence across multiple energy thresholds and intensities. From the figure, it is evident that the P3 flux channel (protons with energies $\geq$10 MeV) consistently exhibits a substantially higher feature importance compared to the other channels (P5 flux channel, $\geq$50 MeV, and P7 flux channel, $\geq$100 MeV). The observed trends of the P3 flux across all scenarios indicate its strong predictive capability and suggest that SEP events are more effectively characterized and identified by the proton flux at lower energy thresholds ($\geq$10 MeV), as expected. 


On the other hand, the flux channels corresponding to higher energies (P5 and P7) show minimal importance, as indicated by the near-zero feature values across. This indicates their limited contribution to model performance, suggesting that protons at higher energy thresholds are less predictive of SEP events in the analyzed scenarios. Additionally, the temporal behavior displayed in the plots highlights that the predictive importance of the P3 flux channel significantly increases as the event onset approaches. This temporal pattern demonstrates the sensitivity of the model to immediate precursors, emphasizing the necessity of shorter lag times or forecasting windows to achieve higher predictive accuracy.



\section{Remarks}

Our experiments demonstrate that temporal proximity to SEP event onset is influential for model accuracy, highlighting the necessity of integrating short-lag observational windows into SEP forecasting models. The observed decrease in predictive skill scores (TSS, HSS, and GSS) at longer lead times underscores the intrinsic challenges associated with early SEP prediction, emphasizing that key precursor signals become increasingly ambiguous as the forecasting window expands. Additionally, the global explainability analysis underscores the importance of proton flux measurements, particularly at the 10 MeV (P3 channel) threshold, in effectively distinguishing and predicting SEP events. This finding suggests potential avenues for future research that may prioritize refining observational and predictive methods that leverage low-energy proton flux measurements. Furthermore, while higher-energy proton flux channels showed limited predictive relevance in this analysis, future studies might explore their role under different modeling frameworks or in combination with other observational parameters to further enhance forecast skills.

Our results clearly indicate that shorter observational lead times significantly improve predictive performance, suggesting a strong dependence of model skill on immediate precursor observations. The proton flux at energies $\geq$10 MeV (P3 channel) was found to be the most influential predictive feature across all scenarios, highlighting its critical importance in forecasting SEP occurrences.

\section{Conclusion}\label{conclusions}

In this study, we proposed and validated a novel approach to enhancing explainability and interpretability in the machine learning-based prediction of SEP events. By integrating global feature mapping into a sliding-window multivariate time series forest model (Slim-TSF), we successfully addressed key limitations associated with the "black-box" nature of traditional ML approaches. Our approach provides practical benefits beyond improved predictive accuracy. By leveraging explainability methods, we facilitated deeper physical insights into model decisions, promoting trust and usability of the forecasts by solar physicists and operational stakeholders. While our results are promising, further exploration of additional solar and heliospheric parameters and the integration of other explainability methods could offer incremental improvements. 

Moreover, integrating explainability techniques such as global feature mapping and counterfactual explanations significantly enhances the interpretability of machine-learning-based forecasts. The application of these methods not only provides critical insights into the underlying predictive relationships but also bridges the gap between purely data-driven predictions and physical understanding. However, challenges persist, notably concerning the uncertainty introduced by small missing intervals at the end of certain observational windows. Additionally, constraints related to data availability, particularly due to the inherent rarity and complexity of SEP events, pose ongoing challenges to achieving robust model generalization across varied solar cycles. Addressing these limitations will require further refinement of interpretability frameworks and continued exploration of techniques to manage class imbalance and incomplete observational data.
Future research directions should include expanding the interpretability framework to incorporate spatially and temporally distributed observations, exploring advanced methods for addressing class imbalance and data scarcity, and developing operationally viable forecasting systems with real-time capabilities. Ultimately, such advancements will contribute significantly to enhancing space weather prediction infrastructure, improving mission planning, and mitigating potential hazards to space-based technologies and astronaut safety.

\section*{Acknowledgment}
This work is supported in part under two grants from NSF (Award \#2104004) and NASA (SWR2O2R Grant \#80NSSC22K0272).

%
%
%
\bibliographystyle{splncs04}
\bibliography{mybib}
\end{document}